\documentclass[letterpaper, 10pt, conference]{ieeeconf}
\IEEEoverridecommandlockouts  
\usepackage{amsmath}
\usepackage{algorithm}
\usepackage{algpseudocode}
\usepackage{amsfonts}
\usepackage[subrefformat=parens,labelformat=parens]{subfig}
\usepackage{graphics} 
\usepackage{graphicx}
\usepackage{amssymb}
\usepackage{caption}
\usepackage{array}
\usepackage{amsmath}
\usepackage{hyperref}
\usepackage{mathtools}
\usepackage{cite}
\usepackage{pgfplots}
\usepackage{tikz}
\usepackage{textcomp}
\usepackage{cleveref}

\usepgfplotslibrary{units}
\usetikzlibrary{pgfplots.units}
\usetikzlibrary{positioning,shadows,shapes,arrows}
\hypersetup{bookmarks=false, hidelinks=true}
\let\norm\undefined 
\DeclarePairedDelimiter\norm{\lVert}{\rVert}

\begin{document}

\title{Autonomous Removal of Perspective Distortion for \\ Robotic Elevator Button Recognition}

\author{Delong Zhu, Jianbang Liu, Nachuan Ma, Zhe Min, and Max Q.-H. Meng
\thanks{The authors are with the Department of Electronic Engineering, The Chinese University of Hong Kong, Shatin, N.T., Hong Kong SAR, China. \textit{email: \{dlzhu, jbliu ncma, zmin, qhmeng\}@ee.cuhk.edu.hk}
}}


\maketitle

\begin{abstract}
Elevator button recognition is considered an indispensable function for enabling the autonomous elevator operation of mobile robots. However, due to unfavorable image conditions and various image distortions, the recognition accuracy remains to be improved. In this paper, we present a novel algorithm that can autonomously correct perspective distortions of elevator panel images. The algorithm first leverages the Gaussian Mixture Model (GMM) to conduct a grid fitting process based on button recognition results, then utilizes the estimated grid centers as reference features to estimate camera motions for correcting perspective distortions. The algorithm performs on a single image autonomously and does not need explicit feature detection or feature matching procedure, which is much more robust to noises and outliers than traditional feature-based geometric approaches. To verify the effectiveness of the algorithm, we collect an elevator panel dataset of 50 images captured from different angles of view. Experimental results show that the proposed algorithm can accurately estimate camera motions and effectively remove perspective distortions. 
\end{abstract}

\vspace{0.5cm}
\section{Introduction}
\label{introduction}
Autonomous elevator operation is regarded as a promising solution to inter-floor navigation of mobile robots, which is typically composed of three components: button recognition, motion planning, and robot control. As the most fundamental step, the performance of button recognition directly determines the success rate and robustness of the whole elevator operation system. 
In order to acquire robust recognition result, traditional approaches prefer leveraging fiducial markers to provide reference anchors, based on which, users have to manually calibrate elevator panel layout and figure out the geometric relationship of each button with respect to the markers. 
Obviously, such system configurations highly depend on human assistance and bring the users much inconvenience. For some applications, e.g., indoor robotic exploration \cite{zhu2018deep, zhu2017hawkeye, semantic2019}, the elevators may not be accessible in advance and hence cannot be calibrated or operated.

To address the problems mentioned above, various button recognition algorithms \cite{zhu2018novel, dong2017autonomous, liu2017recognizing} based on deep learning technique are proposed, which are typically performed on raw images that usually include severe perspective distortions. The key idea of this work is to autonomously remove such distortions to help improve button recognition accuracy. A typical button recognition algorithm usually contains two subtasks: button detection and character recognition \cite{zhu2018novel}, and the proposed distortion removal method can benefit both procedures. Firstly, the layout of elevator buttons usually follows certain patterns, e.g., a regular grid in most cases. Through fitting a grid for the elevator panel, some buttons that are not detected by the detector can be retrieved. Secondly, characters without perspective distortions can be easily identified by the recognizer, meanwhile, the mis-recognized characters can even be corrected by leveraging the rules of button layout \cite{dong2017autonomous}. 
\begin{figure}[t]
	\centering
	\vspace{0.1 cm}
	\includegraphics[width=0.48\textwidth]{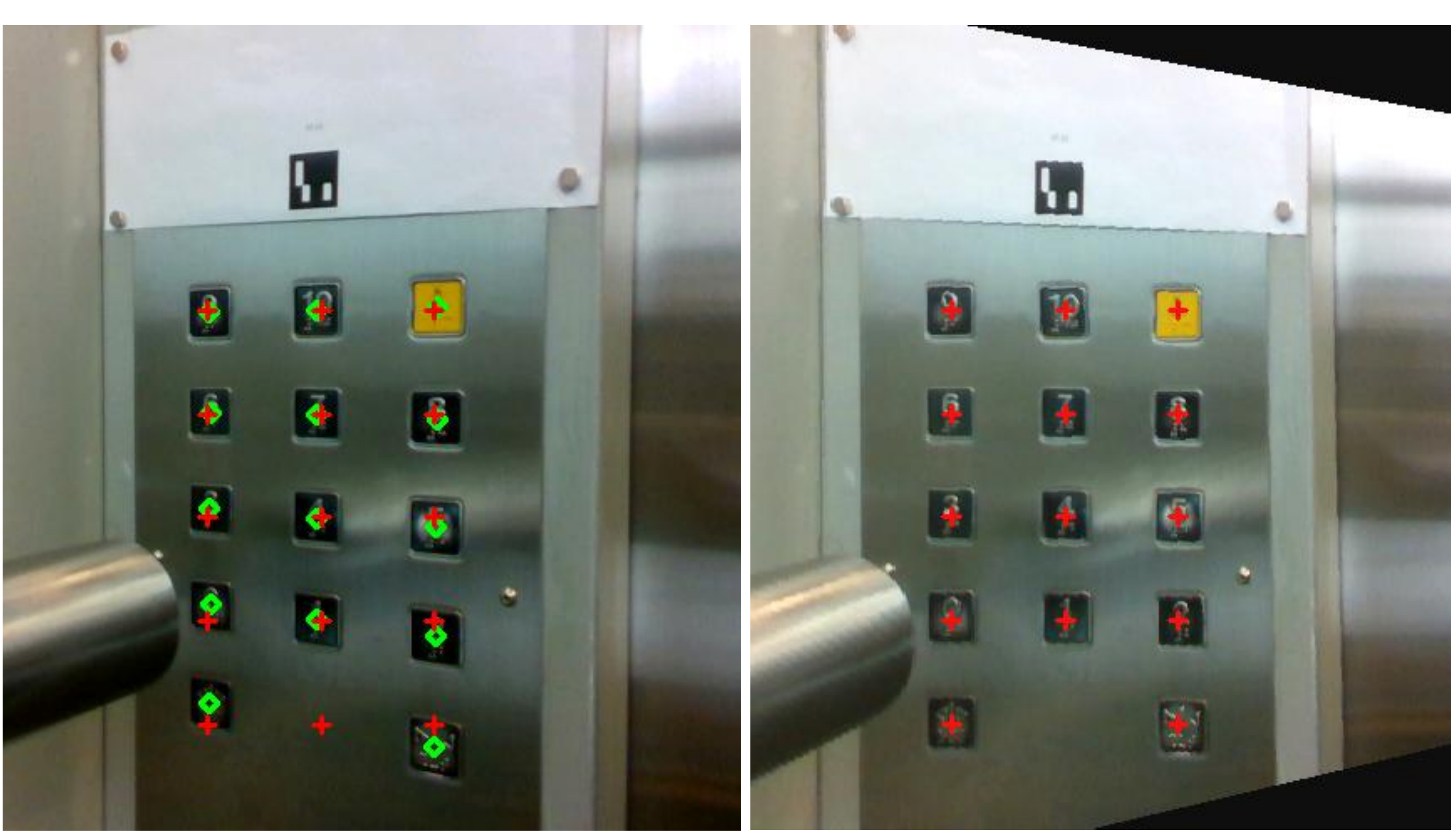}
	\caption{A comparison between original image (left) and corrected image (right). The green diamonds indicate the centers of bounding box predicted by button detector. The red crosses represent the centers of a fitted grid on current image with 5 rows and 3 columns.}
	\label{archt}
\end{figure}

The perspective distortion removal approach consists of two interdependent parts. The first part is a grid fitting algorithm, which models the grid centers as GMM centroids and leverages the Expectation-Maximization (EM) framework to find the optimal centroid positions regarding current detection results. This algorithm is developed based on \cite{Klingbeil2010Autonomous}, in which the images are assumed without distortions. Different from this work, we derive a closed-form solution in this paper and further extend it to handle perspectively distorted images. The second part is a pose estimation algorithm in the normalized image plane. This algorithm takes the optimized grid centers as reference points and calculates the rigid-motion that can align the centers of the predicted bounding boxes and the reference points. Through applying an inverse transformation, the perspective distortion can be removed, based on which, a new round of grid fitting and pose estimation can be performed to achieve more accurate distortion removal. 

The contributions of this paper are summarized as follows:
\begin{itemize}
	\item We propose a perspective distortion removal algorithm that can autonomously estimate distortion parameters using a single elevator panel image.  
	\item We derive a closed-form solution for the maximization step of EM and successfully extend it to handle grid fitting problem on an image with perspective distortions. 
	\item We collect and release an elevator panel dataset with ground-truth poses labeled on every image and initial experimental results are presented as baselines.
\end{itemize}

For the remaining of this paper, we first introduce some related work in Sec. \ref{relate}, then present some preliminary knowledge and the distortion removal algorithm in Sec. \ref{preliminary} and Sec. \ref{method}, respectively. After that, the experimental results and discussions are presented in Sec. \ref{exp}. At the end of this paper, in Sec. \ref{conclusion}, we conclude the paper and present the future work.

\section{Related Work}
\label{relate}
Before the wide application of deep learning techniques in robotics \cite{8665177}, researchers have already attempted to develop button recognition algorithms. Klingbeil \textit{et al.} \cite{Klingbeil2010Autonomous} design a pipeline that can perform button detection and character recognition. With the assumption that all the images are captured without any distortion, this method achieves an accuracy of $86.2\%$ in a testing set with 50 images. Some other approaches \cite{Kim2011Robust} \cite{Wan2014Elevator} are proposed afterwards, but their recognition accuracy and robustness are not satisfying due to the limited capacity of traditional algorithms and the small scale of related datasets. In more recent years, deep learning based large-scale object recognition for robotics becomes a more promising research area. Meanwhile, there emerge an increasing number of publications on the topic of elevator button recognition \cite{zhu2018novel, dong2017autonomous, yang2018intelligent, liu2017recognizing, hsu2018button, wang2018robot}. For instance,
Liu \textit{et al.} \cite{liu2017recognizing} leverage a detection network, the SSD \cite{Liu2016SSD}, to conduct button recognition, which essentially regards the recognition as a multi-object detection problem and hence cannot handle a lot of button categories. To address this problem, Zhu \textit{et al.} \cite{zhu2018novel} propose to integrate a character recognition branch into the Faster-RCNN \cite{Wojna2017Attention} \cite{Ren2015Faster}. In this way, the previous multi-object detection problem is converted to a binary button detection task and a character recognition task, which enables the network to handle hundreds of button classes.

Although there exist various deep learning based button recognition methods, the removal of perspective distortions is seldom mentioned. Nevertheless, such an idea is actually a widely-adopted idea in practice. In \cite{7831854}, the authors utilize a competitive learning method to autonomously remove false positives of pedestrian detectors. In \cite{tglipose}, an encoded pattern is leveraged to help estimate the rigid-body transformations between the target and robot, which is actually a similar idea with this work, except that the reference points are encoded in advance there, whereas we have to estimate these points in this work. From a technical view, the methodologies used in this work share many similarities with point set registration \cite{min2018robust, 8718799, min2019joint, min2016accuracy} and the main difference lies in the parameters that need to be optimized. Point set registration focuses on solving two key problems: point correspondences and parameter optimization \cite{horaud2010rigid}, which are also the cores of grid fitting algorithm.  

\section{Problem Formalization}
\label{preliminary}
Throughout this paper, the following notations are used:
\begin{itemize}
	\item $\mathbf{X}=\left[\mathbf{x}_{1}, \cdots, \mathbf{x}_{N}\right] \in \mathbb{R}^{\,2 \times N}$ - the detected button centers in image plane,
	\item $\hat{\mathbf{X}}=\left[\hat{\mathbf{x}}_{1}, \cdots, \hat{\mathbf{x}}_{N}\right] \in \mathbb{R}^{\,3 \times N}$ - the detected button centers in normalized image plane,
	\item $\mathbf{U}\,=\left[\,\mathbf{u}_{1}, \cdots, \mathbf{u}_{N}\right] \in \mathbb{R}^{2\times N}$ - the centroids of GMM model, which is determined by model parameters $\boldsymbol{\Theta}$,
	\item $\mathbf{z}$ - the hidden variable that adopts one-hot encoding,
	\item $\boldsymbol{o}=[o_x, o_y]^T$ - the origin of the fitted grid, center of the most top-left cell of the grid,
	\item $\Delta=[\Delta_x, \Delta_y]^T$ - the interval distance between grid centers in $x$ and $y$ directions in image plane,
	\item $i, j \in \mathbb{Z}^*$ - the index of grid cells in $x$ and $y$ directions, respectively, and there are $K$ pairs of $(i, j)$,
	\item $\boldsymbol{\Sigma}= \text{diag}([\sigma_x, \sigma_y])$ - the diagonal covariance matrix shared by all the GMM components,
	\item $\boldsymbol{\Theta}\,=(\, \boldsymbol{o}, {\Delta}, \boldsymbol{\Sigma})$ - the parameters of GMM, 
	\item $\boldsymbol{\xi} = [\boldsymbol{\theta}, \boldsymbol{t}]^T$ - the rigid-body motion between bounding box centers and fitted grid centers, 
	\item $\boldsymbol{R}(\boldsymbol{\theta})$ - the matrix representation of angle-axis parameterized rotation $\boldsymbol{\theta}$.
\end{itemize}
 

In this work, the GMM is leveraged to learn the best grid fitting, in which the detected button centers $\mathbf{X}$ is considered to be sampled from a mixture Gaussian as follows: 
\begin{equation}
\begin{aligned}
\mathbf{X} \sim \sum_{k=1}^{K} \pi_k \mathcal{N} (\; \cdot \; |\mathbf{u}_{k}, \boldsymbol{\Sigma}),\\
\end{aligned}
\end{equation}
where $\pi_k$ is the coefficient of GMM and $\sum_{k=1}^{K}  \pi_k  = 1$. The probability of a single sample $\mathbf{x}_{n}$ is calculated as follows: 
\begin{equation}
\label{margin1}
p(\mathbf{x}_{n}|\boldsymbol{\Theta})= \sum_{\mathbf{z}}p(\mathbf{z})p(\mathbf{x}_n|\mathbf{z}) = \sum_{k=1}^{K} \pi_k \mathcal{N} (\mathbf{x}_{n}|\mathbf{u}_{k}, \boldsymbol{\Sigma}),
\end{equation}
where $p(\mathbf{z}) \in \{\pi_1, \cdots, \pi_k \}$ represents a discrete distribution over the hidden variable $\mathbf{z}$, which is a prior that directs the mixture process of Gaussian distributions. Following the conventions in point set registration literature \cite{min2018robust}, $p(\mathbf{z})$ is assumed to be a uniform distribution, i.e. $\pi_k = 1/K$, in this work. 

To account for noises and outliers, an additional uniform distribution is integrated to GMM, 
\begin{equation}
\label{margin2}
p(\mathbf{x}|\boldsymbol{\Theta})=\frac{(1-\alpha)}{C} + \alpha\sum_{k=1}^{K}\frac{1}{K} \mathcal{N} (\mathbf{x}|\mathbf{u}_{k}, \boldsymbol{\Sigma}),
\end{equation}
where $\alpha$ is a balance parameter and we omit the subscript $n$ for simplicity.
In order to make $p(\mathbf{x}_{n})$ a distribution in the image plane, the following constraint should be satisfied,
\begin{equation}
\label{con2}
	\sum_{\mathbf{x}[1]=1}^{w}\sum_{\mathbf{x}[2]=1}^{h} p(\mathbf{x}|\boldsymbol{\Theta}) = 1,
\end{equation}
where the operator $[\, \cdot \,]$ is used to fetch the corresponding elements of vector $\mathbf{x}$.
The solution of Eq. \eqref{margin2} and \eqref{con2} gives us an approximation of $C$, i.e., $C \approx w\times h$, where $w, h$ are the width and height of the image plane, respectively. 

For simplicity, we add an additional dimension to $\mathbf{z}$, thus $ \mathbf{z} \in \mathbb{R}^{\,K+1}$, and then use a more compact formula to denote the marginal distribution presented in Eq. \eqref{margin2},
\begin{equation}
p(\mathbf{x}| \boldsymbol{\Theta})=\sum_{k=1}^{K+1} \pi_k^{\prime} p (\mathbf{x}|\mathbf{z}),
\end{equation}
where $\pi_k^{\prime} = \alpha \pi_k$ for $ 1 \leq k\leq K$ and  $\pi_{K+1}^{\prime} = 1-\alpha$.
The joint distribution of $\mathbf{x}$ and $\mathbf{z}$ also plays an important role in estimating grid fitting parameters. Based on Eq. \eqref{margin1} and \eqref{margin2}, this distribution can be expressed as follows:
\begin{equation}
p(\mathbf{x},\mathbf{z}) = 
\left\{\begin{array}{c}
\begin{aligned}
&\alpha \mathcal{N} (\mathbf{x}|\mathbf{u}_{k}, \boldsymbol{\Sigma}) / K, \;\;\text{if } {\mathbf{z}[k]=1},\\
&(1-\alpha)/ C, \hspace{0.4cm} \text{if }{\mathbf{z}[K+1] = 1}.
\end{aligned}
\end{array}\right.
\end{equation}

Given the predicted button centers $\mathbf{X}$, the optimal $\boldsymbol{\Theta}$ can be achieved by optimizing the following negative log likelihood function:
\begin{equation}
\label{complete}
\begin{aligned}
E(\boldsymbol{\Theta}) &= -\sum_{n=1}^{N} \ln p(\mathbf{x}_{n}|\boldsymbol{\Theta})\\
& = -\sum_{n=1}^{N} \ln \sum_{k=1}^{K+1} \pi_k^{\prime} p (\mathbf{x}|\mathbf{z})\\
& = -\sum_{n=1}^{N} \ln \left[\frac{(1-\alpha)}{C} + \alpha\sum_{k=1}^{K}\frac{1}{K} \mathcal{N} (\mathbf{x}|\mathbf{u}_{k}, \boldsymbol{\Sigma})\right]
\end{aligned}
\end{equation}
where $\mathbf{u}_k = [o_x + \Delta x \cdot i, o_y + \Delta y \cdot j]^T$. Eq. \eqref{complete} is also called \textit{complete log likelihood}, but it is difficult to be directly optimized, as the summation is inside the $\ln$ function. A general framework for solving Eq. \eqref{complete} is the EM algorithm, which essentially utilizes the \textit{expected log likelihood} to iteratively optimize the {complete log likelihood}. 

\section{EM For Distortion Removal}
\label{method}
The EM framework contains two iteratively performed steps: E-step and M-step. In the E-step, we use current parameter values $\boldsymbol{\Theta}^{old}$ to calculate the posterior distribution of latent variable $p(\mathbf{z}|\mathbf{x})$, 
\begin{equation}
\begin{aligned}
\label{posterior}
	p(\mathbf{z}|\mathbf{x}) & = \frac{p(\mathbf{x}, \mathbf{z})}{p(\mathbf{x})} \propto p(\mathbf{x}, \mathbf{z}). \\
\end{aligned}
\end{equation}
Denote a probability distribution table for $p(\mathbf{z}|\mathbf{x})$ and each element of this table records a probability value that is defined as follows:
\begin{equation}
	\gamma(n,k) = p(\mathbf{z}[k]=1|\mathbf{x}_n),
\end{equation}
which actually specifies a type of soft correspondences between detected button centers and the grid centers.

In the M step, we use the posterior distribution to find the expectation of the complete log likelihood, i.e., the expected log likelihood, and maximize this log likelihood function to update the parameter $\boldsymbol{\Theta}$, which can be expressed as follows:
\begin{equation}
	\begin{aligned}
	\boldsymbol{\Theta}^{\text {new}}&=\underset{\boldsymbol{\Theta}}{\arg \max } \mathcal{Q}\left(\boldsymbol{\Theta}, \boldsymbol{\Theta}^{\text {old}}\right) \\
	&=\underset{\boldsymbol{\Theta}}{\arg \max } \sum_{n=1}^{N}\sum_{\mathbf{z}} p(\mathbf{z} | \mathbf{x}_n, \boldsymbol{\theta}^{\text {old}}\,) \ln p(\mathbf{x}_n, \mathbf{z} | \boldsymbol{\Theta}) \\
	& = \underset{\boldsymbol{\Theta}}{\arg \max } \sum_{n=1}^{N}\sum_{\mathbf{z}} p(\mathbf{z} | \mathbf{x}_n, \boldsymbol{\Theta}^{\text {old}}\,)  \ln p(\mathbf{x}_n| \mathbf{z}, \boldsymbol{\Theta}) \\
	& = \underset{\boldsymbol{\Theta}}{\arg \min } \sum_{n=1}^{N} \sum_{k=1}^{K} \boldsymbol{\gamma}_{nk} (\boldsymbol{e}_n^T\mathbf{\Sigma}^{-1}\boldsymbol{e}_n + \ln |{\mathbf{\Sigma}}|), \\
	\end{aligned}
\end{equation}
where $\boldsymbol{e}_n$ is the distance between detected buttons center $\mathbf{x}_n = [x_n, y_n]^T$ and $k^{th}$ grid center indexed by $(i_k, j_k)$,
\begin{equation}
	\boldsymbol{e}_n = \left[
	\begin{matrix}
	x_n - (o_x + \Delta x \cdot i_k) \\
	y_n - (o_y + \Delta y \cdot j_k)
	\end{matrix}
	\right].
\end{equation}
The derivative of $\mathcal{Q}$ w.r.t. $\boldsymbol{\Theta}$ is as follows:
\begin{equation}
\label{derivative}
\begin{aligned}
	\frac{\partial \mathcal{Q}}{\partial \boldsymbol{\Theta}} &= (\frac{\partial \mathcal{Q}}{\partial \boldsymbol{e}})^T	\frac{\partial \boldsymbol{e}}{\partial \boldsymbol{\Theta}} \\
	& = 2\sum_{n,k}\boldsymbol{\gamma}_{nk}(\boldsymbol{\Sigma}^{-1}\boldsymbol{e}_n)^T\frac{\partial \boldsymbol{e}_n}{\partial \boldsymbol{\Theta}},
\end{aligned}
\end{equation}
\begin{equation}
\frac{\partial \boldsymbol{e}_n}{\partial \boldsymbol{\Theta}} =
\left[\begin{array}{cccc}{-1} & 0 & -i_k & 0\\ 
\boldsymbol{0} & {-1} & {0} & -j_k 
\end{array}\right].
\end{equation}
Assume a diagonal covariance matrix and let Eq. \eqref{derivative} be zero, the following equations can be derived,
\begin{equation}
	\begin{aligned}
		\sum_{n,k}\boldsymbol{\gamma}_{nk}\sigma_x[x_n - (o_x + \Delta_x  \cdot i_n)] = 0, \\
		\sum_{n,k}\boldsymbol{\gamma}_{nk}\sigma_y[y_n - (o_y + \Delta_y \cdot j_n)] = 0, \\
		\sum_{n,k}\boldsymbol{\gamma}_{nk}\sigma_x i_n [x_n - (o_x + \Delta_x \cdot i_n)] = 0, \\
		\sum_{n,k}\boldsymbol{\gamma}_{nk}\sigma_y j_n [y_n - (o_y + \Delta_y \cdot j_n)] = 0, \\
	\end{aligned}
\end{equation}
Correspondingly, the closed-form solution of $\boldsymbol{\Theta}$ can be achieved,
\begin{equation}
\begin{aligned}
o_x &= \frac{\sum_{n,k}\boldsymbol{\gamma}_{nk}(x_n - \Delta_x  \cdot i_k)}{\sum_{n,k}\boldsymbol{\gamma}_{nk}}, \\
o_y &= \frac{\sum_{n,k}\boldsymbol{\gamma}_{nk}(y_n - \Delta_y  \cdot j_k)}{\sum_{n,k}\boldsymbol{\gamma}_{nk}}, \\
\Delta_x &= \frac{\sum_{n,k}\boldsymbol{\gamma}_{nk} i_k (x_n - o_x)}{\sum_{n,k}\boldsymbol{\gamma}_{nk}i_k^{2}}, \\
\Delta_x &= \frac{\sum_{n,k}\boldsymbol{\gamma}_{nk} j_k (y_n - o_y)}{\sum_{n,k}\boldsymbol{\gamma}_{nk}j_k^{2}}, \\
\sigma_x & = \frac{1}{N} \sum_{n=1}^{N}\boldsymbol{\gamma}_{nk} \norm{\boldsymbol{e}_n[1]}_2^2 ,\\
\sigma_y & = \frac{1}{N} \sum_{n=1}^{N}\boldsymbol{\gamma}_{nk} \norm{\boldsymbol{e}_n[2]}_2^2 .\\
\end{aligned}
\end{equation}

\begin{figure*}[t]
	\centering
	\includegraphics[width=1.0\textwidth]{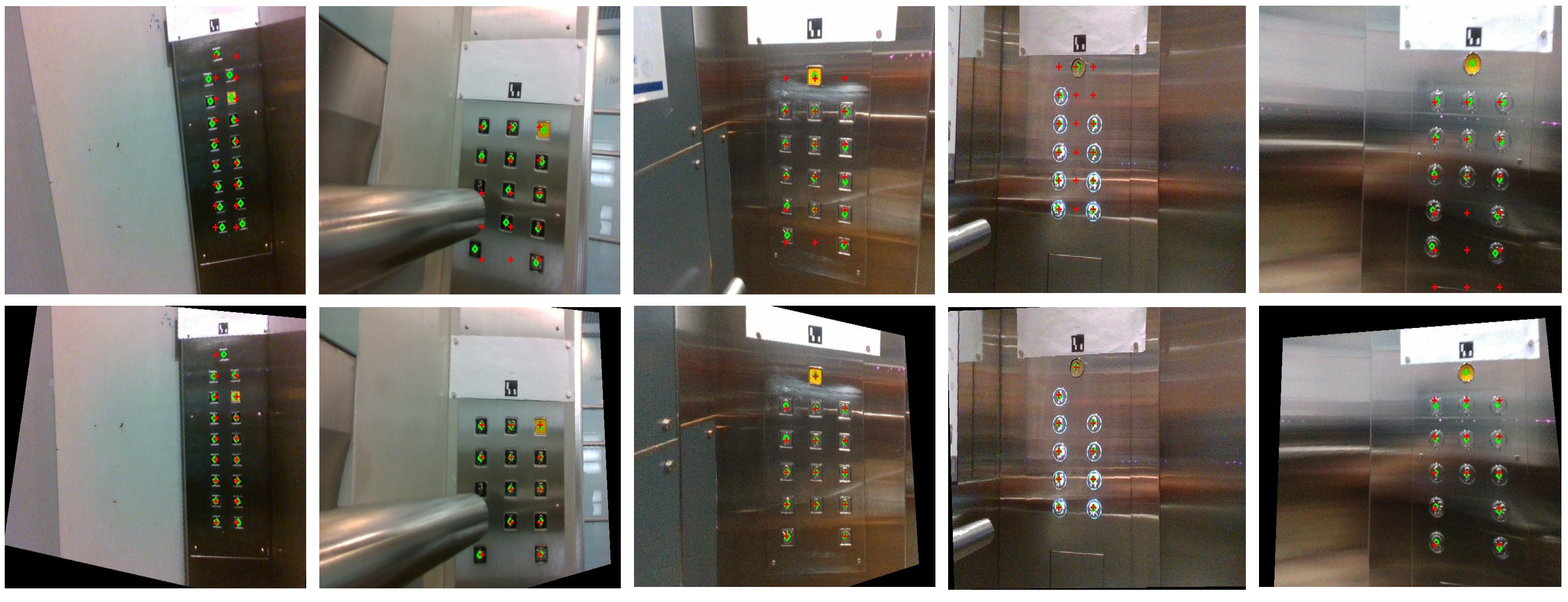}
	\caption{Demonstrations of perspective distortion removal. The first and second row present original and corrected images respectively.}
	\label{recog_demo}
\end{figure*}

Taking advantage of the estimated $\boldsymbol{\Theta}$, we can calculate the reference points $\mathbf{U}$, which is then used for estimation of rigid-motions that lead to perspective distortions. Here, we formalize the pose estimation into a least-square problem as follows:
\begin{equation}
\label{cost_fun}
\begin{aligned}
\boldsymbol{\xi}^{*} &= \arg \min_{\boldsymbol{\xi}} \frac{1}{2} \sum_{n=1}^{N} \boldsymbol{e}_n^T\boldsymbol{e}_n, \\
 \boldsymbol{e}_n &= \hat{\mathbf{x}}_n -\frac{1}{{z}_{n}} (\boldsymbol{R}(\boldsymbol{\theta}) \hat{\mathbf{u}}_n + \boldsymbol{t}),
\end{aligned}
\end{equation}  
where $\boldsymbol{\xi} = [\boldsymbol{\theta}, \boldsymbol{t}]^T$, ${z}_{n}$ is the third dimension of $\boldsymbol{R}(\boldsymbol{\theta}) \hat{\mathbf{u}}_n + \boldsymbol{t}$. The error $\boldsymbol{e}$ is defined in the normalized image plane, which is induced by rigid-body motion $\boldsymbol{\xi}$.

Eq. \eqref{cost_fun} can be efficiently solved using numerical methods, in which the Jacobian matrix is indispensable for performing gradient descent. According to \cite{tim_state}, we directly present the Jacobian matrix of $\boldsymbol{e}$ w.r.t. $\boldsymbol{\xi}$ as follows:    
\begin{equation}
\begin{aligned}
\frac{\partial \boldsymbol{e}}{\partial \boldsymbol{\xi}}=\left[\begin{array}{cccccc}
{\hat{x}\hat{y}} & {-\left(1+\hat{x}^{2}\right)} & {\hat{y}} & {\frac{-1}{z}} & {0} & \frac{\hat{x}}{z}\\
1+\hat{y} & {-\hat{x}\hat{y}} & {-\hat{x}} & {0} & {\frac{-1}{z}} & \frac{\hat{y}}{z}  \end{array}\right].
\end{aligned}
\end{equation}
where $\hat{x}_n$ and $\hat{y}_n$ are the first two dimensions of $(\boldsymbol{R}(\boldsymbol{\theta}) \hat{\mathbf{u}}_n + \boldsymbol{t}) / z_n$. Leveraging the estimated pose $\boldsymbol{R}$ and $\boldsymbol{t}$, we are able to remove the perspective distortion by applying an inverse image warping operation.

\setlength\arrayrulewidth{0.7pt}
\begin{table*}[t]
	\renewcommand{\arraystretch}{1.7}
	\newcolumntype{M}{>{\centering\arraybackslash}m{\dimexpr.10\linewidth-0.8\tabcolsep}}
	\newcolumntype{X}{>{\centering\arraybackslash}m{\dimexpr.13\linewidth-0.8\tabcolsep}}
	\newcolumntype{N}{>{\centering\arraybackslash}m{\dimexpr.10\linewidth-0.8\tabcolsep}}
	\centering
	\caption{Accuracy of distortion removal. In each group, the fist and second row present the image number and accuracy, respectively.}
	\label{control-acc}
	
	\begin{tabular}{MXXXXXN}
		\hline
		{No.} & I-55 & I-660 & I-1180 & I-1610 & I-2190 & Average\\
		1 	  & 0.3026 & 0.0086 & 0.8447 & 2.1007 & 1.1680 & 0.8849\\
		\hline
		{No.} & I-20 & I-450 & I-820 & I-1210 & I-1590 & Average\\
		2 	  & 0.0001 & 1.8100 & 0.0003 & 0.0003 & 0.2041 & 0.4030\\
		\hline
		{No.} & I-20 & I-440 & I-910 & I-1090 & I-1500 & Average\\
		3 	  & 0.0003 & 0.1854 & 0.7971 & 0.2339 & 1.9590 & 0.6351\\
		\hline
		{No.} & I-20 & I-440 & I-840 & I-1000 & I-1320 & Average\\
		4 	  & 0.0000 & 0.0000 & 0.0000 & 0.0332 & 0.3392 & 0.0745\\
		\hline	
		{No.} & I-20 & I-270 & I-410 & I-610 & I-760 & Average\\
		5 	  & 0.5846 & 1.8926 & 0.3684 & 0.8250 & 0.2323 & 0.7806\\
		\hline			
	\end{tabular}
\end{table*}

\section{Experiments}
\label{exp}
To verify the proposed algorithm, we collect a dataset with 25 images from 5 different elevators. All the samples are captured from different angles of views and contain severe perspective distortions. In all the experiments, $\alpha$ is set to 0.8, the initial value of $(o_x, o_y)$ is set to the coordinate of the most top-left detected buttons, $(\Delta_x, \Delta_y)$ is set to (100, 100), and $\sigma_x=640,\, \sigma_y=640$. For pose estimation, the initialization value of $\boldsymbol{\xi}$ is set to (-1.0, -1.0, -1.0, 0.1, 0.1, 0.1). The intrinsic parameter of the camera is 
\begin{equation}
\begin{aligned}
\boldsymbol{K}=\left[\begin{array}{ccc}
320 & 0 & 320 \\
0 & 320 & 240 \\
0 & 0 & 1 \end{array}\right]. 
\end{aligned}
\end{equation}
The detailed experimental results are listed in Table. \ref{control-acc} and some demonstrations are also presented in Fig. \ref{recog_demo}.

To measure the accuracy of distortion removal, the optimal value of Eq. \eqref{cost_fun} is utilized, which indicates an averaged square distance between the reference points and button centers in the corrected normalized image plane. As we can see from Table. \ref{control-acc}, the accuracy value of which is multiplied by 1000, the proposed algorithm achieves very high accuracy on all the five groups of images, which strongly varies the effectiveness of the proposed algorithm. According to Fig. \ref{recog_demo}, we can see that the algorithm is also robust to outliers. Even for panel images that contain irregular button layout, the grid fitting algorithm can also give accurate estimation.

%

\section{Conclusions}
\label{conclusion}
In this paper, we present a novel algorithm for autonomously removing perspective distortions on a single image. The algorithm takes as input the outcomes of button recognizer, then conducts grid fitting and pose estimation procedures, and finally outputs the corrected images, which can help improve the accuracy of character recognition. Currently, the algorithm can only handle internal panel images that contain a number of buttons. For external elevator panel images, as there are not enough button samples,  the algorithm may easily fail. In the next step, we will integrate more feature points, e.g., button corners, into the framework to make it capable of handling external elevator panels. We will also collect a larger dataset to verify the robustness of the algorithm, based on which further improvement will be made.

\section*{Acknowledgment}
This project is partially supported by the Hong Kong RGC GRF grants \#14200618 and Shenzhen Science and Technology Innovation projects c.02.17.00601 awarded to Max Q.-H. Meng.



\bibliographystyle{IEEEtran}
\bibliography{ROBIO}



\end{document}